\pdfoutput=1

\documentclass[11pt]{article}

\usepackage[final]{acl} 

\usepackage{times}
\usepackage{latexsym}
\usepackage{amssymb}
\usepackage{pifont}
\usepackage{booktabs}%

\usepackage{etaremune}
\usepackage[normalem]{ulem}

\usepackage[T1]{fontenc}

\usepackage[utf8]{inputenc}

\usepackage{microtype}

\usepackage{inconsolata}

\usepackage{graphicx}

\usepackage{enumitem}

\usepackage{tabu}
\usepackage{multirow}
\usepackage{booktabs}
\usepackage{arydshln}
\usepackage{xspace}
\usepackage{amsmath}

\usepackage{soul}     
\usepackage{xcolor}   
\setstcolor{red}      

\begin{document}

\author{
  Guilin Hu\textsuperscript{*} \quad
  Malek Itani\textsuperscript{*} \quad
  Tuochao Chen \quad
  Shyamnath Gollakota \\
  \textsuperscript{*}Co-primary student authors\\
  \textsuperscript{1}Paul G. Allen School of Computer Science \& Engineering, University of Washington \\
  \texttt{\{guilinhu,malek,tuochao,gshyam\}@cs.washington.edu}
}

\title{Proactive Hearing Assistants that Isolate Egocentric Conversations}




\newcommand{\squishlist}{%
  \begin{itemize}[itemsep=1pt,parsep=2pt,topsep=3pt,partopsep=0pt,
                  leftmargin=0em,itemindent=1em,labelwidth=1em,labelsep=0.5em]%
}
\newcommand{\squishend}{%
  \end{itemize}%
}



\makeatletter
\renewcommand{\sectionautorefname}{\S\@gobble}
\renewcommand{\subsectionautorefname}{\S\@gobble}
\renewcommand{\subsubsectionautorefname}{\S\@gobble}
\renewcommand{\appendixautorefname}{Appendix \@gobble}
\makeatother

\pagenumbering{gobble}

\newcommand{\xref}[1]{\S\ref{#1}}

\maketitle

\begin{abstract}

We introduce proactive hearing assistants\footnote{Our hearing  assistants are `proactive' in that they  infer and adapt to conversational engagement without user commands.} that automatically identify and separate the wearer's conversation partners, without requiring explicit prompts. Our system operates on egocentric binaural audio and uses the wearer’s self-speech as an anchor, leveraging turn-taking behavior and dialogue dynamics to infer conversational partners and suppress others. To enable real-time, on-device operation, we propose a dual-model architecture: a lightweight streaming model runs every 12.5~ms for low-latency extraction of the conversation partners, while a slower model runs less frequently to capture longer-range conversational dynamics. 
Results on real-world 2- and 3-speaker conversation test sets, collected with binaural egocentric hardware from 11 participants totaling 6.8 hours, show  generalization in identifying and isolating conversational partners in multi-conversation settings. Our work marks a step toward hearing assistants that  adapt proactively to conversational dynamics and engagement. More information can be found on our website:  \url{https://proactivehearing.cs.washington.edu/}


\end{abstract}

\section{Introduction}

Human hearing is remarkably adaptable, yet fundamentally limited in crowded auditory environments. In such settings, isolating relevant voices, known as the cocktail party problem, becomes especially difficult. For individuals with hearing loss, distinguishing overlapping conversations can result in cognitive overload and listening fatigue~\cite{hearingaidscocktail}.


Existing hearing assistants, like augmented devices, wireless earbuds and hearing aids, are ``reactive'' in that users manually prompt the devices to pick specific sound sources via spatial filtering or phone-based interfaces~\cite{semantichearing,tsh}. However, these approaches struggle in multi-party conversations where speakers are spatially dispersed or involve more than two speakers, making manual enrollment impractical.

We propose an alternative: real-time proactive hearing assistants that automatically identify and enhance the voices involved in a conversation with the wearer, without explicit prompts. Our system processes egocentric binaural audio to dynamically track conversational partners and suppress others, adapting to engagement naturally and seamlessly, without explicit user commands or prompts.

\begin{figure}[t!]
    \centering
    \vskip -0.1in
    \includegraphics[width=\linewidth]{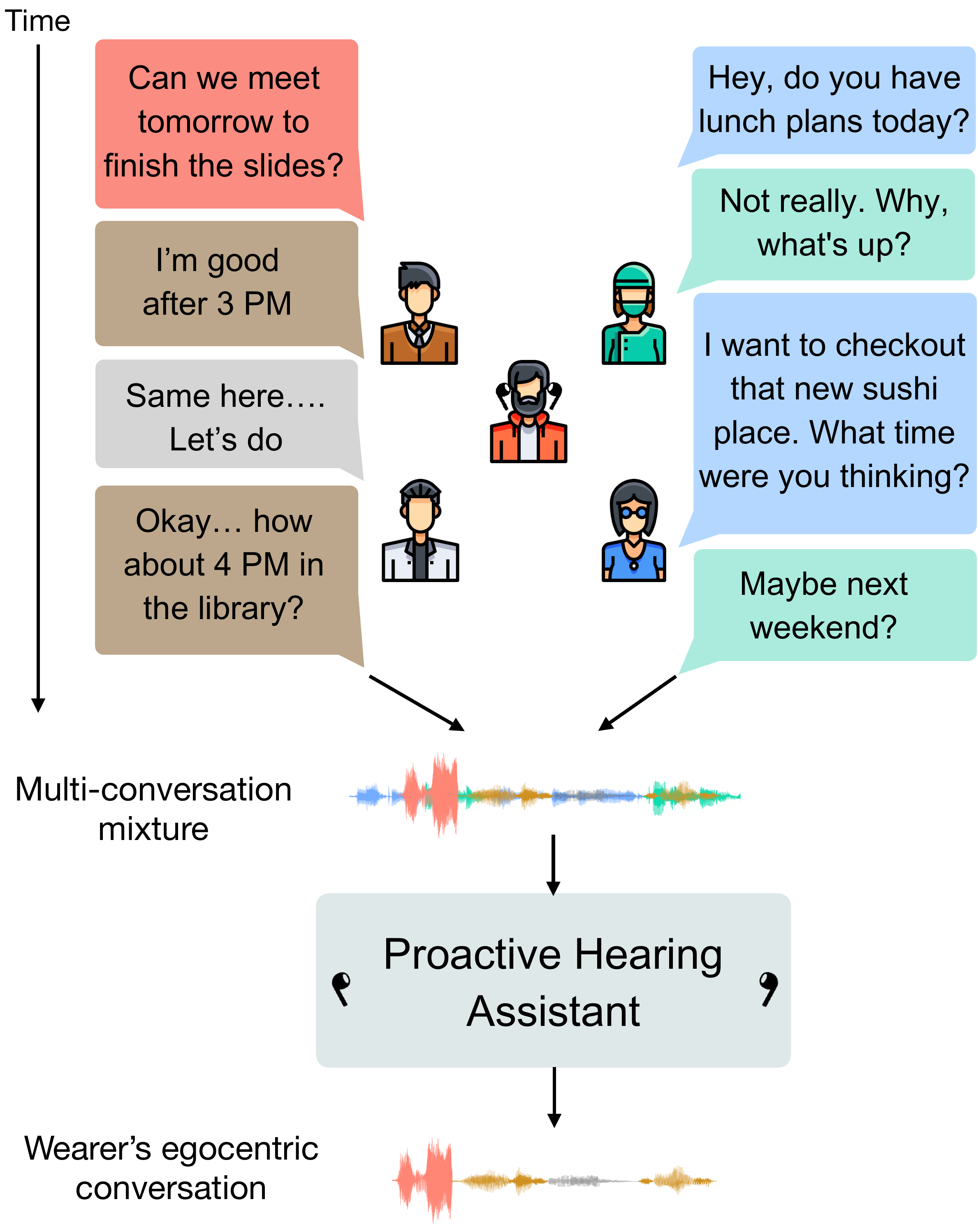}
    \vskip -0.1in
    \caption{In multi-conversation settings, our proactive hearing assistant uses conversational turn-taking  dynamics to automatically infers the wearer's conversation  partners and suppresses others in real-time. }
    \vskip -0.15in
    \label{fig:teaser}
\end{figure}

This task poses three key challenges:
(1) identifying and separating conversational partners,
(2) operating on-device in real-time with low latency, and
(3) generalizing to real-world, egocentric, multi-party  environments.

Our approach builds on insights from core NLP tasks like turn-taking prediction, speaker diarization, and dialog modeling, to design a proactive hearing assistant. We make three key contributions:
(1) an anchoring mechanism based on the wearer’s self-speech to track conversation partners,
(2) a dual-model architecture that enables low-latency, real-time processing, and
(3) a real-world end-to-end evaluation using egocentric binaural conversational recordings captured with wearable hardware.

Concretely, we anchor the system on the wearer’s self-speech, extracted using a beamformer trained on egocentric audio. The assistant activates when the wearer speaks  for a few seconds, signaling conversational intent. The assistant leverages  turn-taking cues, such as alternating speech, low overlap, and temporal coordination, to identify conversational partners. These interactional patterns, well-studied in dialogue systems~\cite{turntake2,frontiers,tce}, allow the proactive assistant to infer engagement in real time and selectively separate the voices of relevant speakers.

To meet real-time constraints, the system processes audio in short streaming chunks. However, as conversations unfold, the sequence length grows, increasing the memory demands of attention-based models. Since full self-attention scales quadratically with sequence length~\cite{ainslie2020etcencodinglongstructured,10889061}, achieving both long-context awareness and low-latency performance requires a carefully designed architecture.

To balance real-time responsiveness and conversational context length, we use a dual-model architecture:  A fast streaming model runs every 12.5~ms, extracting the target conversation in real time. A slower  model runs once per second  and provides periodic longer-term conversational embeddings, capturing conversation turn-taking and discourse structure without incurring full-attention memory.

We train on diverse speech and conversational English and Mandarin datasets, including  Candor~\cite{candor}, LibriTTS~\cite{zen2019libritts} and RAMC~\cite{ramc}, spatialized to emulate egocentric conditions. We evaluate our models on out-of-distribution  SpokenWOZ~\cite{si2023spokenwoz} and the Japanese Duplex Conversation Dataset \cite{JapaneseDuplex2025}. We also collect real-world  2- and 3-speaker conversational testset using binaural egocentric hardware from 11 participants, totaling 6.8 hours.

In both out-of-distribution and real-world  egocentric  settings, our system accurately identifies conversational partners, with accuracies and confusion rates of 80-92\% and 1.5-2.2\% respectively. It also improves speech quality of the conversation partners by 7.22-11.95~dB (SISDRi), and operates in real time on embedded and mobile devices.

This work shows a path toward proactive hearing assistants that go beyond source separation to infer who the user wants to hear, adapting to  conversation dynamics in a way that aligns closely with goals in dialogue systems, speech understanding, and human-AI interaction.  It also offers future potential for adapting LLM agents to track spoken conversations in noisy, multi-party settings.

\section{Related work}

\noindent\textbf{Conversational dynamics.} 
Understanding multi-party conversation structure has long been a focus in dialogue systems and speech processing. Prior work has explored speech recognition~\cite{asr4conversation}, speaker diarization~\cite{diarization4conversation}, and speech-driven question answering~\cite{qa4conversation}, often under idealized conditions without interfering speakers. Dialogue-level sentiment analysis and discourse segmentation have also been explored in clean settings~\cite{sa4conversation, sdu4conversation}.

Turn-taking is a central feature of conversational dynamics~\cite{frontiers}, and has been studied using corpus-based models~\cite{turntake1, turntake2, turntake3} that identify patterns such as alternating speech, pauses, and backchannels. Recent approaches model turn-taking directly~\cite{turn_model,  turn_predict,nguyen-etal-2025-spirit}, including listener behavior prediction in dyadic settings~\cite{9879511}.

Most relevant to our task is Target Conversation Extraction (TCE)~\cite{tce}, which uses turn-taking cues to extract a target conversation. Our work differs in three key aspects: (1) TCE operates offline and requires future context, making it unsuitable for real-time use; (2) it relies on explicit speaker embeddings, while we use self-speech extracted from egocentric binaural audio as a natural anchor; and (3) it uses monaural recordings, whereas we focus on realistic, spatialized egocentric audio from wearable devices.

\vskip 0.02in\noindent{\bf Audiovisual speech understanding.}   Our work intersects with research in Active Speaker Detection (ASD) and Active Speaker Localization (ASL). ASD systems identify who is speaking using audio-visual correlations or facial features~\cite{1544886}, while ASL  focuses on spatial localization \cite{8578556, 9879100, easycom}. Recent work in Selective Auditory Attention Localization extends this to inferring whom the user is attending to, using egocentric video and audio~\cite{10203787, 10447323}.

Efforts in egocentric video understanding have explored detecting social engagement~\cite{6247805} and speaker attention~\cite{ego4d}. For instance, the Ego4D benchmark includes a ``Talking to Me'' task focused on identifying who is addressing the camera wearer. However, these tasks typically stop at detection. In contrast, we go further: identifying, separating, and enhancing all speakers engaged with the wearer, in real time and on-device, under real-world constraints.

\vskip 0.02in\noindent{\bf Auditory attention decoding.}  
 Research in this domain  attempts to infer the target speaker by correlating brain activity (e.g., EEG or fNIRS) with competing audio streams~\cite{aad1, Choudhari2024.02.05.579018,aad2}. While promising, these systems lack real-time deployment capabilities and require bulky or invasive hardware. Even with miniaturized in-ear EEG sensors~\cite{cEEGrid,berkeleyearable}, challenges remain in noisy, real-world settings with multiple speakers~\cite{separationeeg}. In contrast, our work explores an dialog-based approach that aligns better with practical hearing assistance, leveraging self-speech as an implicit signal of attention and engagement.

\vskip 0.02in\noindent{\bf Proactive assistants.} 
 Prior work has explored proactive interaction in task planning~\cite{zhang-etal-2024-ask}, user modeling~\cite{lu2024proactiveagentshiftingllm}, and conversational guidance~\cite{llamapie}. However, these systems focus on information-seeking or planning tasks, which are complementary to our task.

\vskip 0.02in\noindent{\bf Augmented hearing.} 
Contemporary hearing systems support  selection of target sound sources, e.g., a speaker or sound class, via  spatial filtering or manual enrollment~\cite{semantichearing, tsh,soundbubble,knowledgeboosting}. Apple’s Conversation Awareness mode~\cite{apple-conversationawareness} reduces background volume upon detecting wearer speech, but does not perform speaker separation or conversational tracking. 


\section{Proactive Hearing Assistants}

\begin{figure*}[t]
    \centering
    \includegraphics[width=1\linewidth]{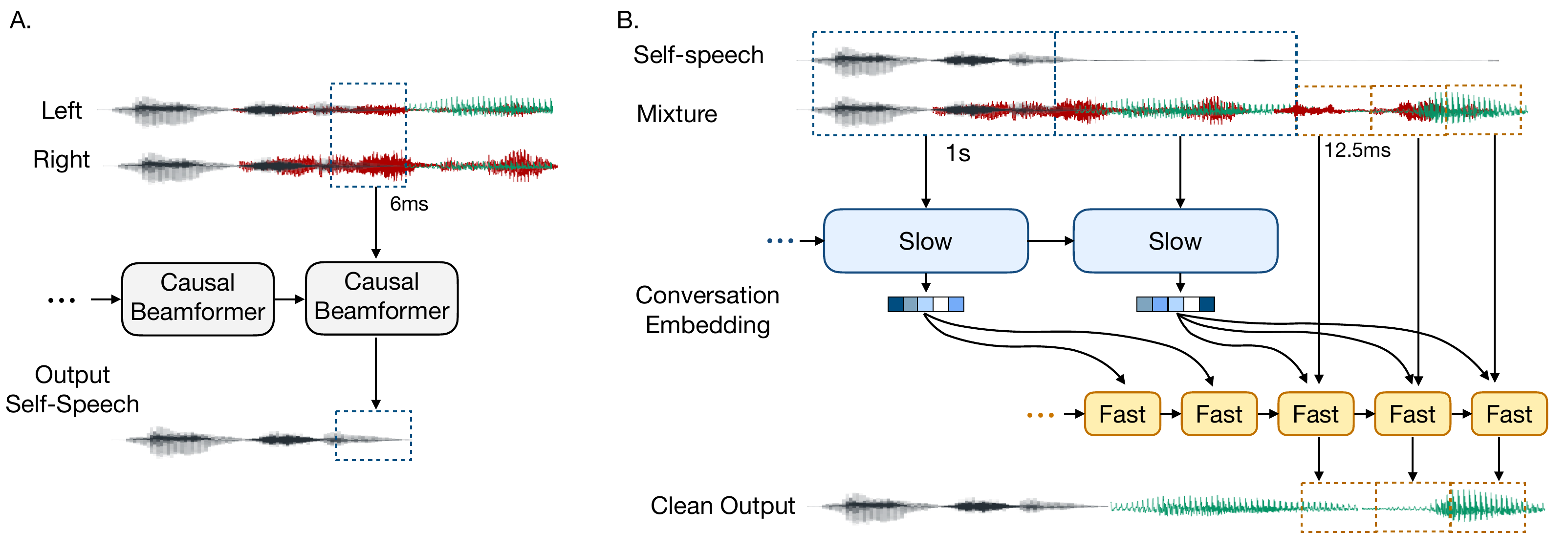}
    \vskip -0.14in
    \caption{Overview of our model pipeline.     A. The streaming beamformer extracts the wearer's self-speech from the binaural mixture. B. Dual-model architecture: the slow model runs every 1s ($T$) on the mixture and self-speech to produce a conversation embedding; the fast model runs every 12.5~ms ($\tau$) on the current mixture and  embedding from the previous 1s ($T$), to output the cleaned target conversation.
    }
    \vskip -0.15in
    \label{fig:pipeline}
\end{figure*}


\subsection{Problem formulation}
\label{sec:problem_formulation}

The input egocentric audio stream can be decomposed into three components: the target conversation involving the wearer, interfering conversations, and background noise. 


The target conversation  consists of the wearer's self-speech and the speech of their conversational partners. Notably, in egocentric recordings, the wearer's own speech is typically  louder  than all other voices.

Our goal is to identify and separate the wearer's conversation partners, which can be done by isolating the target conversation and suppressing model output during the wearer’s own speech, using on-device voice activity detection (e.g., as in AirPods~\cite{apple-conversationawareness}). The system can then  output the conversation partners' speech into the ear.

The system must handle dynamic conversational settings, where speakers may join or leave the conversation at any time, following natural turn-taking patterns. Real-world dialogue includes backchannels and overlaps, requiring the model to adapt as speakers shift between target and interfering conversations. For example, a speaker might begin as part of the target conversation (e.g., at a dinner table) but later engage in a separate, interfering conversation, requiring the system to adapt.

Once the conversation partners are extracted, their voices must be rendered to the wearer with minimal delay to preserve a natural conversational experience. Thus, the system must process audio in small chunks of 10–20~ms to maintain a  latency below the perceptual threshold. Real-time operation requires each chunk to be processed faster than it is recorded. Because offloading to a phone or cloud introduces communication delays (10–30~ms over Bluetooth and 100–200~ms over the Internet),  streaming processing must occur on-device on compute-limited embedded platforms.

An assumption we make is that the wearer is an active participant in the conversation. Thus, passive listening, such as eavesdropping, is a non-goal. 

\subsection{Proactive assistant modeling}\label{sec:modeling}
Fig.~\ref{fig:pipeline} shows the full  architecture, including all sub-networks and the data pipeline. The beamformer and the slow conversational embedding model operate on audio chunks of $T$ second length, while the fast streaming model runs on much shorter chunks of $\tau$ seconds ($\tau \ll T$). The beamformer takes the egocentric binaural audio stream and isolates the wearer's own voice by beamforming toward their mouth. This self-speech, along with the monaural egocentric audio mixture, is inputted to the slow  model, which generates an embedding every $T$ seconds. This embedding is then used by the fast streaming model to guide target conversation extraction for upcoming audio chunks. The fast model receives  a single egocentric audio stream and the conversational embedding as input.

\subsubsection{Dual-model processing}\label{sec:dual}
Conversations can occur continuously for very long durations. Thus, it is useful to utilize attention to effectively model and retain minutes-long contextual sequences of audio. However, with attention, it is challenging to meet the strict real-time requirements needed for proactive hearing assistants on hardware with tight processing capabilities. 

Specifically, as conversation length increases, so does the number of chunks, leading to longer input sequences for the attention mechanism. This is problematic, as the memory requirements of full self-attention scale quadratically with sequence length~\cite{ainslie2020etcencodinglongstructured,10889061}. Ideally, we want to maintain long-context awareness for accurate filtering while ensuring low memory usage and real-time performance. 

We employ a dual-model pipeline. It incorporates a high-latency, attention-based network to model long  sequences and extract a  conversation embedding, and a low-latency, low-complexity LSTM-based network that integrates this conversation embedding to estimate the target conversation in small chunks. Since the fast model does not directly attend to historical context, its memory footprint is low. Further, because the large model processes fewer, longer chunks, it attends over fewer tokens for the same conversation duration, enabling it to efficiently capture extended context (details in~\xref{sec:dualmodeldetails}).

Several key design choices support real-time performance. First, the beamformed self-speech is not fed into the fast model, as doing so would introduce additional processing latency that would violate real-time constraints. Second, the  slow embedding model processes audio $T$ seconds behind the fast stream. This decoupling allows these models to run remotely on say a smartphone. Further, it prevents their higher processing latency from affecting streaming performance, but introduces a tradeoff: larger $T$ values reduce the system’s responsiveness to  conversational dynamics (see~\xref{sec:ablation}).  Third, both the fast and slow models use monaural rather than binaural audio as input. This reduces computational load on the fast model and ensures the models focus on conversational turn-taking and dynamics rather than spatial cues.

\subsection{Training strategy}\label{sec:training}

Our models must generalize to egocentric binaural conversations with 2–3 participants and handle dynamic scenarios where participants may leave the target conversation and join an interfering one. To jointly model conversation tracking and source separation, we require mixtures of a target egocentric conversation with a separate interfering conversation with no shared speakers. However, since the wearer's self-speech dominates in egocentric audio, we cannot simply mix two egocentric recordings. Instead, we need passive third-person binaural recordings to construct realistic mixtures.

Existing egocentric datasets like EgoCom~\cite{egocom} and EasyCom~\cite{easycom} are unsuitable for this purpose:  EgoCom features the same host in all recordings and  both datasets   lack third-person binaural recordings needed for  mixture synthesis.

Instead, we use non-egocentric datasets and spatialize them to simulate egocentric scenarios. We train on the Candor dataset~\cite{candor}, which contains 850 hours of high-quality 2-speaker English conversations, and RAMC~\cite{ramc}, which has 180 hours of 2-speaker Mandarin conversations. Both provide clean audio, speaker IDs, and timestamps. Large open-source datasets with 3-speaker conversations or complex dynamics (e.g., speakers switching conversations) are scarce.

\subsubsection{Synthetic dataset creation}
To address this, we adopt the time-preserving method from~\cite{tce} and generate {five} synthetic datasets (see~\xref{sec:datasetgen}):
\squishlist
    \item {\it Libri (2spk).} We align LibriTTS~\cite{zen2019libritts} audio from two random English speakers with RAMC~\cite{ramc} 2-spk timestamps, replacing the original Mandarin utterances.
    
    \item {\it Libri (3spk).} With RAMC timestamps, we randomly assign each turn to one of three LibriTTS speakers, creating a synthetic 3-spk conversation.
    
    \item {\it Libri (leaving).} A speaker active in the first 20 seconds of the 3-spk conversation leaves and reappears in the interfering conversation between 20–40 seconds, simulating speaker dynamics.

    \item {\it Libri (4spk) and (5spk) (Evaluation only).} {Two test-only datasets where RAMC test set timestamps are used to generate synthetic four- and five-speaker conversations by randomly assigning each turn to one of four or five LibriTTS speakers.}

\squishend

\subsubsection{Training procedure}\label{sec:trainingrecipe}

We generate  mixtures by combining a target conversation with an interfering conversation and  noise. Each target conversation starts at least 5 seconds of the wearer’s self-speech, so the models can anchor to the wearer. Training proceeds in three stages.

 We pretrain on the training splits of the three synthetic datasets and Candor mixtures. The fast streaming model and the slow conversational embedding model are trained jointly, with a negative SNR loss computed on the fast  model’s output to reconstruct the target conversation. The conversational embedding model receives ground-truth self-speech as input.

In the second stage, to simulate egocentric hearing, we spatialize the synthetic and Candor datasets (see~\xref{sec:spatialize}). Ground-truth self-speech is replaced with the output of a pretrained beamformer, which serves as input to the slow  model. Both models are trained jointly using the same loss function.

To address the distribution shift between Candor (Zoom-based, first-time interactions) and real-world, in-person conversations between familiar participants, in the final stage, we finetune the model by perturbing the amount of silence and overlap between speaker utterances (see~\xref{sec:trainingdetails})).

\subsection{On-device real-time inference}
The fast streaming model runs on a low-power embedded device, while the slower conversational embedding  model can operate remotely on device with more compute. 
To meet real-time requirements, we run the fast streaming model on an embedded Orange Pi 5B and the slower conversational model on Apple M2 silicon, supported by commodity wearable devices. The fast model processes 12.5~ms audio chunks in 8.9~ms on average, while the slow model processes 1-second chunks in 41.3~ms. {In addition, we profiled memory usage for the slow and fast models. We run streaming inference of the slow and fast models for 100 runs. Then we measure the peak memory usage averaged over 100 runs. Peak memory is 591.47 MB (slow model) and 86.33 MB (fast model) during streaming inference.}

\section{Evaluation}


\subsection{Metrics}\label{sec:metrics}

Since the beamformer already outputs self-speech and the proactive assistant aims to help the wearer hear conversational partners, we compute four metrics for the {\it partners' speech} segments output by the models (see~\xref{sec:beamformerdetails}  for self-speech results).

\squishlist
\item {\it SISDRi:} Scale-Invariant Signal-to-Distortion Ratio improvement (SISDRi) quantifies how much the target speech is enhanced relative to the noisy input. Higher values indicate better separation and preservation of the target speech.

\item {\it $\Delta$PESQ: } Perceptual Evaluation of Speech Quality (PESQ) estimates speech quality based on human auditory perception. $\Delta$PESQ measures the perceptual improvement over the input mixture.

\item {\it Accuracy (Acc):} Measures how often we correctly select the conversational partner at each conversation turn. A correct selection occurs when: (1) the conversational partner's  SISDRi > 0, and (2) it exceeds all interfering speakers' SISDRi.


\item {\it Confusion Rate (CR):}  How often we select an interfering speaker over the target. This occurs when: (1) the interfering speaker's SISDRi > 0, and (2) it exceeds the conversational partner's SISDRi.

\squishend


\begin{table*}[t!]
\caption{Evaluation on English (Libri, Candor, SpokenWoZ) and Mandarin (RAMC) testsets.}
\vskip -0.1in
\centering
  \setlength{\tabcolsep}{5.3pt}
{\footnotesize
\begin{tabular}{lcccccccc}
\toprule
& \multicolumn{4}{c}{Non-spatialized} & \multicolumn{4}{c}{Spatialized} \\
\cmidrule(r){2-5} \cmidrule(l){6-9}
\textbf{Metrics} & SISDRi($\uparrow$) & Acc ($\uparrow$) & CR ($\downarrow$) & $\Delta$PESQ($\uparrow$) & SISDRi($\uparrow$) & Acc($\uparrow$) & CR($\downarrow$)  & $\Delta$PESQ($\uparrow$) \\
\midrule
\textbf{Baseline Model (SE)} &&&&&&&&\\
\midrule
Synthetic Libri & -1.95 (1.94) & 26.8\% & 26.3\% & -0.16 (0.14) & -3.31 (2.23) & 10.5\% & 27.7\% & -0.13 (0.13)\\
Candor  & -5.16 (3.19) & 13.5\% & 24.4\% & -0.29 (0.22) & -3.05 (2.84) & 18.8\% & 21.7\% & -0.16 (0.17)\\
SpokenWoz (OOD) & -5.05 (4.04) & 14.7\% & 14.7\% & -0.48 (0.33) & -3.45 (3.78) & 16.8\% & 9.2\% & -0.16 (0.21)\\
RAMC  & -4.28 (6.01) & 13.6\% & 27.9\% & -0.32 (0.28) & -4.30 (3.37) & 6.8\% & 19.9\% & -0.17 (0.17)\\
\midrule
\textbf{Our Dual Models} &&&&&&\\
\midrule
Synthetic Libri & 11.70 (4.56) & 96.3\% & 1.4\% & 1.11 (0.30) & 14.62 (6.05) & 96.8\% & 0.8\% & 0.63 (0.23) \\
Candor  &  6.75 (4.29) & 87.0\% & 2.7\% & 0.64 (0.31) & 9.82 (3.77) & 93.9\% & 1.1\% & 0.56 (0.20)\\
SpokenWoz (OOD)  & 7.27 (6.11) & 84.5\% & 4.3\% & 0.58 (0.44) &11.95 (6.22) & 92.1\% & 1.5\% & 0.52 (0.25) \\
RAMC  & 6.50 (7.45) & 85.5\% & 5.6\% & 0.63 (0.41) & 8.05 (9.29) & 78.0\% & 9.4\% & 0.26 (0.40) \\
\midrule
Libri (2spk) & 12.48 (4.28) & 98.4\% & 0.6\% & 1.14 (0.24) & 15.68 (5.46) & 99.2\% & 0.2\% &  0.65 (0.22) \\
Libri (3spk)  & 12.03 (4.00) & 96.6\% & 1.2\% & 1.16 (0.25) & 14.29 (6.08) & 96.1\% & 0.6\% & 0.63 (0.21) \\
Libri (leaving) & 10.58 (5.10) & 94.1\% & 2.5\% & 1.03 (0.36) & 13.89 (6.42) & 95.7\% & 1.5\% & 0.60 (0.24)\\
\bottomrule
\end{tabular}
}
\label{tab:big_results}
\vskip -0.15in
\end{table*}


\subsection{Testsets}\label{sec:testsets}
We evaluate our models on several test sets: synthetic 2-speaker conversations (Libri 2spk), 3-speaker conversations (Libri 3spk), speaker-switching conversations (Libri leaving), and the Candor test set. There is no speaker {or turn-taking timestamp} overlap between the training, validation, and test sets, ensuring that the models have not seen the test conversations or speakers during training.


We also assess generalization by testing the English-trained models on both the RAMC Mandarin test set, which contains no turn-taking timestamp data from training, {and the Japanese Duplex Conversation Dataset \cite{JapaneseDuplex2025}. We also tested the model on Libri (4 spk) and Libri (5 spk), where the model was not trained on such a large number of speakers.} Finally, we evaluate on the out-of-distribution SpokenWOZ~\cite{si2023spokenwoz} 2-speaker conversation dataset. Since SpokenWOZ contains relatively short utterances, we do not enforce the condition that the wearer speaks for at least 5 consecutive seconds at the beginning. Instead, we randomly select two recordings with disjoint speakers and designate the first speaker in the target conversation as the wearer.

\subsection{Results}\label{sec:results}
Table~\ref{tab:big_results} shows evaluation results on several open-source conversational datasets. We use DeepFilterNet2~\cite{schroter2022deepfilternet2}, a widely adopted speech enhancement model, as our baseline. In the non-spatialized setting, our dual-model consistently outperforms the baseline across all four metrics. On the synthetic Libri conversational dataset, the model achieves significant improvements in both SISDR and PESQ under various conditions, including 2-, 3-speaker and speaker-leaving scenarios. Additionally, our model achieves a high accuracy to pick the conversational partners and  a very low confusion rate to pick the interfering speakers.  In contrast, the baseline model enhances  speech uniformly without distinguishing between target and interfering speakers. As it is not conversation-aware or capable of speech separation, it fails to deliver SISDR improvements.

\begin{figure}[t]
    \centering
    \vskip -0.1in
    \includegraphics[width=0.6\linewidth]{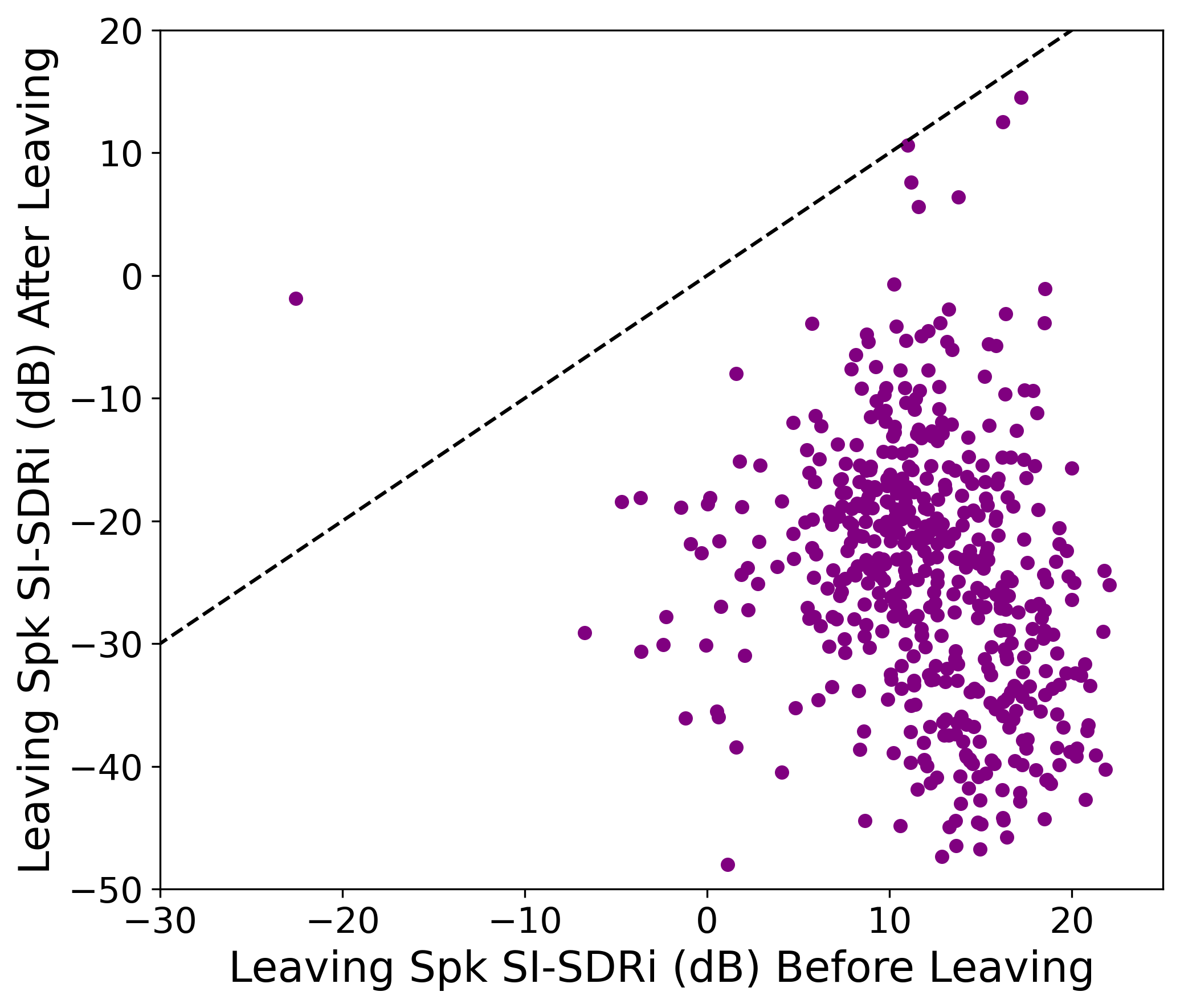}
    \vskip -0.1in
    \caption{Model enhances then suppresses speaker following  shift from target to interfering conversation.}
    \vskip -0.15in
    \label{fig:leaving_scatter}
\end{figure}

Fig.~\ref{fig:leaving_scatter} shows a scatterplot from the Libri (leaving) test set, where a speaker transitions from the target conversation to the interfering one. The plot depicts the SISDRi achieved by our dual-model for this speaker, both before and after leaving the target conversation. While part of the target conversation, the speaker receives a positive SISDRi, indicating successful enhancement. After switching to the interfering conversation, the SISDRi becomes negative, showing that the model correctly suppresses the speaker once they are no longer part of the conversational flow. This shows  the model’s ability to adapt to dynamic, multi-party interactions.

\begin{table}[t!]
\caption{SISDRi results for generalization.}
\vskip -0.1in
\centering
  \setlength{\tabcolsep}{5pt} 
{\footnotesize
\begin{tabular}{lc}
\toprule
\textbf{Dataset} & SISDRi ($\uparrow$) \\
\midrule
Libri (2spk)        & 12.48 (4.28) \\
Libri (3spk)        & 12.03 (4.00) \\
Libri (4spk)        & 11.94 (4.46) \\
Libri (5spk)        & 11.85 (4.58) \\
\midrule
RAMC                & 6.50 (7.45)  \\
Japanese (OOD)      & 7.92 (5.19)  \\
\bottomrule
\end{tabular}
}
\label{tab:sisdr_datasets}
\vskip -0.15in
\end{table}

{To evaluate the model’s ability to generalize to conversations involving more than three speakers, we constructed Libri 4- and 5-speaker datasets, which were only used as test sets. As shown in Table \ref{tab:sisdr_datasets}, although the model was not trained on conversations with such a number of speakers, it achieved performance comparable to that observed on the Libri 2- and 3-speaker datasets. This suggests the model generalizes well to conversations with previously unseen numbers of target speakers.}

In Table~\ref{tab:big_results}, we further evaluate the model on SpokenWOZ, an out-of-distribution (OOD) English dataset, highlighting the generalization ability of both the model and training approach.  {In addition, to assess the model’s ability to generalize across languages, we evaluate it on the Mandarin RAMC dataset and the Japanese Duplex Conversation Dataset \cite{JapaneseDuplex2025}. As shown in Table \ref{tab:sisdr_datasets}, our model reaches a 6.5 dB and a 7.92 dB SISDRi, respectively. This shows that even though our model is trained solely on English speakers, it can generalize  to conversations in other languages, because it is primarily learning the turn-taking patterns. 

We also evaluate our model on the noisy Libri (2-spk) test set, where WHAM! noise from its test split is added. The model achieves an SISDRi of 10.37 dB, demonstrating its ability to generalize to noisy conditions despite not being trained on such data. Further fine-tuning on noisy data for 35 epochs using the WHAM! training split yields an improved SISDRi of 11.84 dB.

Finally, we  evaluate all models on spatialized test sets that emulate egocentric  conditions, as shown in Table~\ref{tab:big_results}. In egocentric scenarios, speech from other speakers tends to have lower amplitude than the wearer’s self-speech due to physical distance. As a result, after spatializing the synthetic Libri conversational dataset, the average input SISDR for the conversation partners drops from 1 dB to –10 dB, and the average input PESQ decreases from 2.52 to 2.04.
Given this challenging setting, our model achieves a 14.62 dB improvement in SISDR and a 0.63 increase in PESQ, while maintaining high speaker selection accuracy (96.8\%) and a low confusion rate (0.8\%). We also observe similarly strong performance across other  datasets including spatialized Candor and on the out-of-distribution spatialized SpokenWOZ testset. 


\begin{table}[t!]
\caption{Subjective evaluation results (5-point scale).}
\vskip -0.1in
\centering
  \setlength{\tabcolsep}{6pt} 
{\footnotesize
\begin{tabular}{lcc}
\toprule
\textbf{Question} & \textbf{Model Output} & \textbf{Mixture} \\
\midrule
Noise suppression          & 4.29 (1.19) & 1.67 (1.04) \\
Comprehension & 4.35 (1.02) & 1.97 (0.93) \\
Effort                     & 4.45 (0.95) & 1.97 (0.96) \\
Overall MOS                & 4.30 (1.14) & 1.88 (1.02) \\
\bottomrule
\end{tabular}
}
\label{tab:user_study_table}
\vskip -0.15in
\end{table}

\subsection{Subjective human evaluation}

{To evaluate the model from a user-centric perspective, we conducted a user study with 11 participants (8 males, 3 females, 1 non-binary) with an age range of 21-65. Each listened to six random conversations from the Candor dataset, experiencing both the original mixture and the model output in a random order. Following \cite{10.1145/3613904.3642057}, we asked participants four 5-point scale questions about their experience in focusing on the target conversation (see \xref{sec:user_study}). As shown in Table \ref{tab:user_study_table}, the proposed system improves user-perceived quality across all four aspects, raising the overall mean opinion score from 1.88 to 4.30.}




\subsection{Ablation studies}\label{sec:ablation}

\noindent\textit{Dual-model versus single model.} We compare our dual-model approach with a single fast streaming model that uses the self-speech and mixture audio as input. The single model achieves an SISDR improvement of only 1.45 dB for the conversation partners, much lower than the dual model's 12.48 dB. This shows that without the support of the slower conversation embedding model, the fast model alone, struggles to capture conversational dynamics effectively. 


\vskip 0.02in\noindent{\it Update rate for conversational  embeddings.} Since the fast streaming model relies on conversation embeddings generated by the slow model, we compare two embedding update intervals: 1 second and 4 seconds. We train the dual models on the {Libri 2-speaker  training set for ten epochs each and evaluate them on the corresponding test set. Increasing the update interval from 1s to 4s leads to a drop of 1.22~dB in SISDRi. 


\begin{table}[t!]
\caption{Evaluation on real egocentric conversations.}
\vskip -0.1in
\centering
  \setlength{\tabcolsep}{5.3pt}
{\footnotesize
\begin{tabular}{lccc}
\toprule
\textbf{Metrics} & SISDRi($\uparrow$) & Acc ($\uparrow$) & CR ($\downarrow$)  \\
\midrule
\textbf{Number of speakers} & & &\\
\midrule
    2 speakers & 7.84 (6.79) & 85.0\% & 1.1\% \\
    3 speakers & 6.00 (7.14) & 73.4\% & 3.7\% \\
\midrule 
\textbf{Augmentation} &  &  & \\
\midrule 
  \ding{55} & 5.49 (5.30) & 77.9\% & 1.2\% \\
  \ding{51} & 7.22 (6.96) & 80.0\% & 2.2\% \\
\bottomrule
\end{tabular}
}
\label{tab:hardware_result}
\vskip -0.15in
\end{table}

\vskip 0.02in\noindent{\it Speaker embedding versus self-speech.} Instead of anchoring conversation extraction on self-speech, we also explore using the wearer's speaker embedding~\cite{variani2014deep}. Following~\cite{tce}, we compute 256-dimensional d-vectors from clean wearer speech, and provide these as embeddings to the slow model. Using speaker embeddings reduces the  SISDRi by 2.65~dB  compared to self-speech, likely due to temporal variability in speech characteristics  and lossy representation,   which reduce  embedding reliability.


\vskip 0.02in\noindent{\it Beamforming versus self-speech}. 
To study the impact of using the beamformer's self-speech output versus the ground truth self-speech from the conversation mixture, we use the model trained on spatialized data from stage 2 and evaluate it on Libri (2-spk)  test set. The difference in SISDRi between the two modes was less than 0.38~dB.}

\begin{table}[t!]
\caption{Impact of perturbing the turn-taking in human conversations.  (SD=standard deviation)}
\vskip -0.1in
\centering
  \setlength{\tabcolsep}{6pt} 
{\footnotesize
\begin{tabular}{lc}
\toprule
\textbf{Perturbation SD} & \textbf{SISDRi (dB)} \\ 
\midrule
No perturbation & 6.75 \\ 
0.5s            & 6.25 \\
1s              & 5.68 \\
1.5s            & 5.25 \\
2s              & 4.79 \\
2.5s            & 4.50 \\
3s              & 4.16 \\
\bottomrule
\end{tabular}
}
\label{tab:perturbation}
\vskip -0.15in
\end{table}

\vskip 0.02in\noindent{\it Impact of turn-taking disruption.} {We performed an ablation on the Candor test set to assess the impact of turn-taking pattern disruption. By perturbing inter-utterance silence durations with shifts sampled from a normal distribution with mean 0 and varying standard deviation (SD), we increasingly disrupted the natural turn-taking structure. Table.  \ref{tab:perturbation} shows that as SD and overlap ratio increases, performance gradually degrades, as this breaks the turn-taking structure that the model leverages to separate the targets.} 

\vskip 0.02in\noindent{\it Context length.} {We trained models with different context lengths (full context, 10s, 5s, 1s) by masking the slow model’s self-attention to limit each token's access to past tokens on Libri 2spk training set. We then evaluated on the Libri 2spk testset. Compared to the model with full context access, SISDRi dropped by 2.12 dB, 4.06 dB, and 5.74 dB for context lengths of 10s, 5s, and 1s, respectively. This demonstrates that access to long-term context is a key factor for the system's performance.}

\section{Real-World Egocentric Recordings}

We recruited 11 participants (2 female, 9 male) with an  age range of 21--39. The dataset comprises a total of  6.8 hours of binaural egocentric audio recordings, including seven two-speaker and five three-speaker conversations, each lasting approximately 10 minutes. The participants engaged in open-ended discussions in English, on topics such as food, hobbies, recent activities, research, workouts, and travel plans, with no constraints on subject matter. All recordings took place in an environment with typical background noise, including HVAC and ambient sounds. A summary of the conversation statistics is provided in Table~\ref{tab:realworldstats}.

All sessions took place in the same  acoustic environment so they can be mixed for creating mixtures. During recording, each speaker wore   a pair of binaural microphones (Sonic Presence SP15C) and connected it to a smartphone to  capture egocentric audio of the conversation. Further, in each conversation, a silent participant served as a listener by wearing the microphones and standing in the vicinity of the speakers.   These passive recordings, which lack dominant self-speech, serve as representative samples of interfering conversations.

\begin{figure}[t!]
    \centering
    \includegraphics[width=0.65\linewidth]{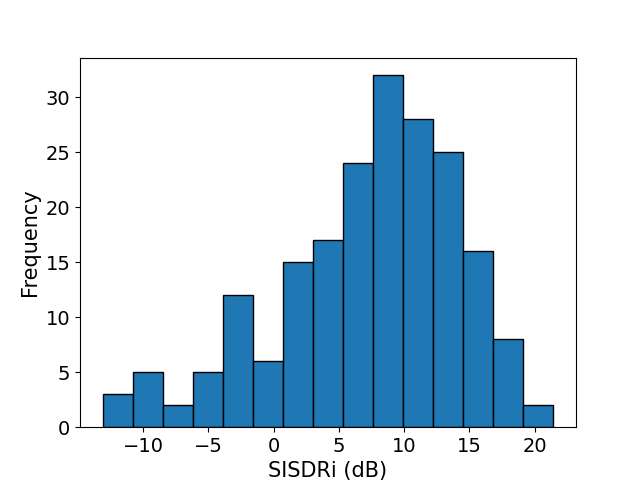}
    \vskip -0.1in
    \caption{SISDRi histogram on egocentric recordings.}
    \vskip -0.15in
    \label{fig:hist}
\end{figure}

\subsection{Data pre-processing}

Since each conversation participant  recorded their own egocentric audio, their self-speech appears with the highest amplitude in their recordings. Using these recordings alongside our beamformer network, we estimated the speech activity timestamps for each speaker. The authors manually verified these timestamps to ensure their quality.

Conversation mixtures were created by combining audio from a target speaker with that of a listener in a separate interfering conversation. To avoid amplifying noise, denoising (Sainburg et al., 2020; Sainburg, 2019) was applied only to the target audio, while interfering audio remained unprocessed to preserve realistic ambient noise. Speakers were not shared across the two conversations. Interfering conversations always involved two speakers, while target conversations had two or three. Each sample was constructed to begin with at least 3 seconds of self-speech, and none from a conversation partner. Input SNRs for the target conversation  were uniformly sampled between –10 and 10 dB. We generated 200 conversation mixtures to serve as out-of-distribution test set for our model.

\begin{table}[t!]
\caption{When the conversation partners start speaking, how quickly does the model pick them up?}
\vskip -0.1in
  \setlength{\tabcolsep}{3.8pt}
\centering
{\footnotesize
\begin{tabular}{lccccc}
\toprule
\textbf{Chunk} & 0-2s & 2-4s & 4-6s  & 6-8s & 8-10s  \\
\midrule
SISDRi (dB) & 4.77 & 8.04 &  8.17 & 8.69 & 9.16 \\
\bottomrule
\end{tabular}
}
\label{tab:pickup}
\vskip -0.05in
\end{table}

\begin{table}[t!]
\caption{Effects of turn-change gap between target conversation and interfering conversation.}
\vskip -0.1in
  \setlength{\tabcolsep}{3.8pt}
\centering
{\footnotesize
\begin{tabular}{lccccc}
\toprule
\textbf{Turn-change} & 0-1s & 1-2s & 2-4s & 4-6s & > 6s  \\
\textbf{Gap} &  & & & &  \\
\midrule
Proportion & 11.3\% & 12.2\%  & 20.7\% & 15.5\%  & 40.3\% \\
SISDRi (dB) & 4.98 & 8.03  & 7.80 & 8.22  & 8.46 \\
\bottomrule
\end{tabular}
}
\label{tab:collision}
\vskip -0.15in
\end{table}

\subsection{Real-world Results}

Table~\ref{tab:hardware_result} shows performance on real-world egocentric recordings with 2- and 3-speaker conversation mixtures.  The performance drop with 3-speaker target conversations is because the three speakers turn-taking dynamic in our training data is all synthesized. Fine-tuning on real 3-speaker conversation datasets may further improve results. 

These results show  real-world generalization from simulated training data. Around 80\% of conversations have a positive SISDRi (Fig.~\ref{fig:hist}).  These results also show the benefit of augmentation described  in~\xref{sec:trainingdetails}, which improves performance by 1.73 dB by creating a more diverse distribution of overlaps and silence in the training set. 


{\it How quickly is a conversation partner picked up?} We investigate how fast the model enhances a conversation partner after they begin talking. Using a 2-second sliding window over each non-wearer turn in the real-world egocentric dataset, Table~\ref{tab:pickup} shows the average SISDRi,  averaged over all turns and samples in the real-world egocentric dataset. In the first 0–2 seconds, SISDRi is 4.77 dB, indicating initial adaptation. After 2 seconds, it exceeds 8 dB, showing the model quickly adapts to  conversational partners within a turn.


{\it How does turn-change collision impact performance?}  We examine how overlapping turn transitions in target and interfering conversations affect performance. For each self-to-other turn change in the target conversation, we compute the time gap to the nearest turn change in the interference. A small gap indicates simultaneous speaker transitions. Table~\ref{tab:pickup} shows that 11.3\% of turns have gaps under 1 second, where SISDRi drops to 4.98 dB. This suggests that closely timed turn-changes can confuse the model. Future work could address this by incorporating conversation content.

{\it What happens with extended periods of wearer silence?} 
 Fig.~\ref{fig:self_speech_absent} shows a real-world example where the wearer did not speak for over 2 minutes. The purple curve indicates the SISDRi of the conversational partner in 30-second windows; grey areas show when the wearer was speaking. SISDRi stayed above 5~dB during intermittent speech but dropped below zero during prolonged silence, indicating the model failed. Performance recovered once the wearer resumed speaking, highlighting the model’s reliance on self-speech as an anchor.

\begin{figure}[t!]
    \centering
    \includegraphics[width=0.75\linewidth]{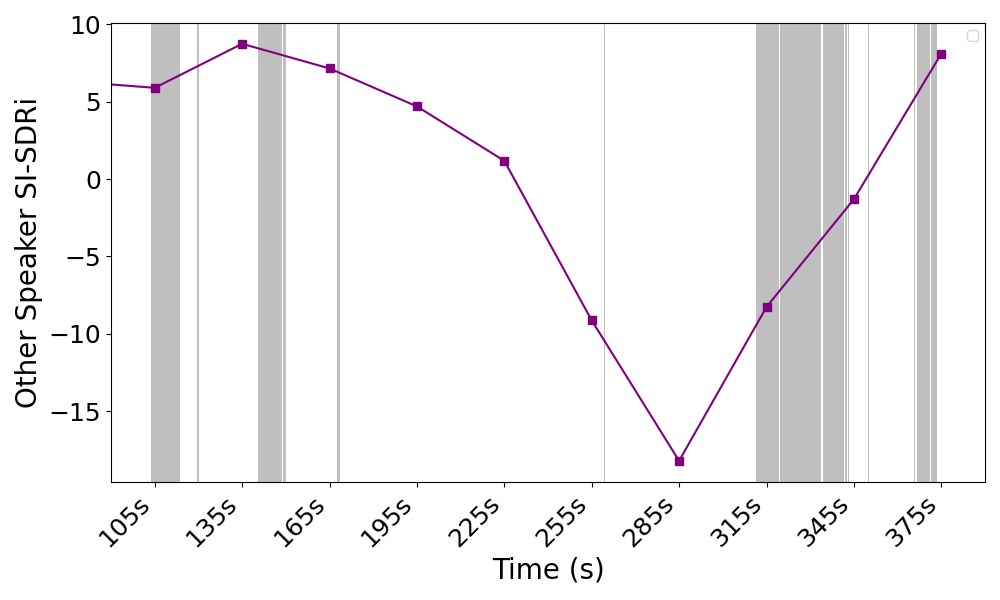}
    \vskip -0.1in
    \caption{Extended periods of wearer silence. The gray regions denote durations were the  wearer was active.}
    \vskip -0.15in
    \label{fig:self_speech_absent}
\end{figure}

\section{Conclusion}

 We present the first real-time, proactive hearing assistant that automatically identifies the wearer’s conversational partners and suppresses unrelated speech, without requiring explicit user prompts. Our system runs on-device and  generalizes to real-world egocentric recordings despite being trained only on synthetic data. By leveraging turn-taking cues to model conversational engagement, our approach connects speech separation with core dialogue modeling tasks. This work takes an important step towards proactive hearing assistants that interpret and adapt to  conversation dynamics.

\section{Limitations and risks}

{\bf Limitations.} Our system is designed for scenarios in which the wearer is an active participant in a conversation, using self-speech as an anchor to identify conversational partners. It is not suited for passive listening, such as eavesdropping or passive consumption. 

The current implementation prioritizes real-time, on-device performance and  incorporates conversational turn-taking. While this design choice supports low-latency operation, it may limit the system's ability to disambiguate overlapping speakers, especially when multiple speakers begin speaking simultaneously. Incorporating lightweight content-aware models could be a direction for future work.


In addition, while the model generalizes to real-world egocentric  recordings without fine-tuning on such data, performance could likely benefit from supervised adaptation to real-world {acoustic and conversation} conditions.

{Finally, although the model achieved cross-linguistic generalization in evaluations on English, Mandarin, and Japanese datasets}, cultural and linguistic differences in turn-taking behavior~\cite{turntake2} suggest that further fine-tuning for language- or culture-specific dynamics may improve robustness.

\vskip 0.05in\noindent{\bf Ethical considerations. } 
Proactive hearing assistants hold promise for improving communication access for individuals with hearing loss, particularly in dynamic and crowded settings. They may be especially valuable for older adults or users with limited dexterity, for whom manual control interfaces may be impractical.

However, there are important risks. Incorrect speaker detection may suppress relevant voices or amplify unrelated ones. Such errors are particularly concerning in high-stakes or fast-moving conversational contexts. Improving this remains a key area for future work.

Additionally, if the assistant fails or behaves unpredictably, users should have a clear and intuitive means to override or adjust system behavior. One practical solution could be a physical control (e.g., a tactile button) to temporarily disable the assistant or reset its state. Addressing these through transparent design, user-centric controls, and robust real-world evaluation will be essential for safe and responsible deployment.

\bibliography{paper}

\appendix

\section{Dual Model architecture details}\label{sec:dualmodeldetails}

As shown in Fig.~\ref{fig:pipeline}, our architecture includes a fast streaming model and a slower conversation embedding model. The streaming model outputs audio with minimal latency, processing each chunk as it arrives. The slow model buffers $T$ seconds of audio to capture long-term conversational dynamics and generates a conversation embedding, which conditions the streaming model for the next $T$ seconds before being updated.

The conversation embedding model also takes the wearer's self-speech as input, estimated using a neural beamformer. While the beamformer adds some latency, it is negligible compared to $T$ and does not affect streaming model latency. The self-speech is concatenated with the noisy audio along the channel dimension and passed to the conversation embedding model.

Both the streaming and conversation embedding models are based on TF-GridNet~\cite{tfgridnet} and operate on audio in the time-frequency (TF) domain. We first convert time-domain audio signal $x \in \mathbf{R}^{C \times t}$, where $C$ is the number of channels and $t$ is the number of frames, using the short-time Fourier Transform (STFT) to obtain the TF-representation $X \in \mathbf{C}^{C \times F \times L}$, where $F$ is the number of frequency bins, and $L = \frac{t}{\tau}$ is the number of time steps after STFT. The real and imaginary components are concatenated along the channel dimension and the resulting tensor $X' \in \mathbf{R}^{2C \times F \times L}$ is provided as the input. 

The conversation embedding model first maps $X'$ to a $D$-channel latent space using a $3\times3$ 2D causal convolutional layer to get $Z_e \in \mathbf{R}^{D \times F \times L}$. Then, the input is processed by a stack of six extraction blocks, each of which consists of a local module and a global module. The local module processes audio information within a $T$ second chunk. It uses bidirectional LSTMs to 1) model the spectral information within the same time step, and 2) model the temporal information within the same frequency bin over exactly $T$ second chunks. This latter process requires that the model wait for $T$ seconds before it can process the sequence of chunks. The global module models relationships across sequences of $T$ second chunks. Specifically, we average pool the information from every $T$ seconds to reduce the temporal resolution and use self-attention on this pooled representation. To ensure causality, attention weights are masked using a lower-triangular matrix, allowing each time step to attend only to previous steps. We use 4 attention heads and absolute positional encoding. Following the global module, we replicate every time step in the pooled representation to retrieve a tensor with the original number of timesteps before pooling. After the last extraction block, we simulate the slow model's algorithmic latency by shifting the result backwards in time by $T$ seconds, inserting zeros at the beginning. This time-varying conversation embedding $E \in \mathbf{R}^{D \times F \times L }$ is returned and can be used to condition the streaming model.

The streaming model also maps $X'$ to a latent representation $Z_s \in \mathbf{R}^{D \times F \times L}$ using a $3\times3$ 2D causal convolutional layer and processes the resulting tensor through six extraction blocks. The model is conditioned on the conversation embedding by multiplying it, element-wise, with the feature map between the first and second extraction blocks. The extraction block uses a bidirectional LSTM to model sequences of frequencies within the same time frame, but replaces the bidirectional temporal LSTM with a unidirectional LSTM to reduce latency and discards the global module entirely. After the last extraction block, we use a deconvolution layer to convert the data back to the TF-domain $Y'\in \mathbf{R}^{2C \times T \times F}$. Finally, we use an inverse STFT and overlap-add to reconstruct the output time-domain signal $x \in \mathbf{R}^{1 \times T}$


We adopt the dual-window method for time–frequency transformation from~\cite{low_latency_stft}. Using this framework, we use an STFT with a chunk size of 200 samples (12.5~ms) and a lookback and lookahead of 32 samples (2~ms). The output window size for the inverse STFT is 232 samples, i.e. we discard the first 32 samples of the inverse FFT output every STFT frame. We use rectangular synthesis and analysis windows. Both models use a latent dimension $D=32$, and an LSTM hidden dimension $H=32$. The local modules of the embedding module use unfolding to reduce the number of steps to process in the time and frequency sequences. This unfolding operation has a kernel size of 2 and a stride size of 2. The global modules project the tensor onto a smaller subspace with only 2 channels before applying self-attention.

The  conversation embedding model has 986K parameters and the streaming model has 491K parameters.

\section{Beamformer model   details}\label{sec:beamformerdetails}

Our beamformer model follows the architecture in~\cite{soundbubble}, excluding the frequency compression modules. To minimize algorithmic latency, we once again use the dual-window method for time–frequency transformation from~\cite{low_latency_stft}. We use a chunk size of 96 samples (6~ms), with a lookback of 96 samples (4~ms) and a lookahead of 64 samples (6~ms). The encoder consists of 3×3 2D causal convolution layers, producing a 32-dimensional latent representation. The model then processes the input with 6 GridNet blocks and LSTMs with a hidden dimension of 32. The inverse DFT uses a 160-sample output window, discarding the first 96 samples during overlap-add. The network outputs two channels, which are averaged to produce the final single-channel beamformer output.


\subsection{Beamformer datasets }\label{sec:beamformerdatasets}

The beamformer is a neural network designed to extract the user's self-speech in the presence of surrounding speech and noise. It takes binaural audio recorded from a headset worn by the user as input. Because the network relies on spatial cues, such as inter-channel phase and level differences, it is especially sensitive to spatial features that are difficult to model accurately in simulated environments.

To address this, we first pretrain the beamformer on a large dataset of synthetically generated binaural recordings, then finetune it on a smaller set of real-world binaural recordings. The final model is a lightweight beamformer with 174K parameters that generalizes well to real-world acoustic conditions, making it well-suited for use as a self-speech extractor for our real-time hearing assistant.


To train the beamformer on synthetic data, we create a dataset of 5-second audio mixtures. Each mixture includes speech from a user wearing a binaural headset and 1 to 5 interfering speakers, sampled with equal probability. All speech signals are drawn from LibriSpeech. If an audio clip exceeds 5 seconds, we randomly crop a 5-second segment; if it is shorter, we pad it with a random duration of silence. Simulated egocentric binaural signals are generated using the method described in~\xref{sec:spatialize}, and these signals are summed to form the final mixture. Interfering signals are scaled so that the mixture’s SNR is uniformly distributed between -5 and 20dB. Training and validation audio are sampled from LibriSpeech’s \verb+train-clean-360+ and \verb+dev-clean+ splits, respectively. The final synthetic dataset contains 20K training mixtures and 1K val mixtures.


We further train the beamformer using real-world data. For this, we collected 3 hours of self-speech from 9 participants across 15 different rooms, along with 4 hours of interfering speech from 4 participants in 3 rooms. To generate training examples, we create 5-second binaural mixtures by combining a 5-second self-speech clip with 0 to 5 interfering speech clips of the same length. Each 5-second clip is formed by extracting 2–5 seconds of active speech from a speaker and padding it with a random amount of silence.

All audio clips are scaled so that their power (in dBFS) follows a normal distribution with a mean of -25 and a standard deviation of 5. Additionally, we include a 5-second binaural noise clip from the binaural WHAM! dataset. The WHAM! noise is randomly scaled by a factor in $[0,1]$ before being added to the mixture. Noise clips for training and validation are drawn from the \verb+tr+ and \verb+cv+ splits of the WHAM! dataset, respectively. The final mixture is obtained by summing the self-speech, interfering speech, and noise. Interfering speech and noise are scaled to produce an overall SNR uniformly distributed in $[-5,20]$~dB. These real-world mixtures are generated on the fly during training, and we use 1,000 mixtures for validation.


\subsection{Beamformer training }\label{sec:beamformertrainingdetails}

The beamformer is trained in two stages: (1) on synthetic data, and (2) fine-tuned on real-world recordings. In both stages, we use a batch size of 8, apply gradient clipping with a max norm of 0.1, and optimize using AdamW \cite{adamw} with a weight decay of 0.01.

The synthetic data training stage is trained for 200 epochs on  negative SNR Loss. We vary the learning rate based on a schedule. For the first 10 epochs, we linearly increase the learning rate from 0.0001 to 0.001. Then, we maintain this learning rate for 140 epochs. Finally, we further train for 50 epochs, halving the learning rate every 15 epochs.

The real world data fine-tuning stage occurs over 300 epochs, with each epoch defined as 20K iterations. Here, we use the following composite loss function:  $L(\hat{x}, x) = 10 ||x-\hat{x}||_1 + L_{MR}(\hat{x}, x)$, where $x$ is the target signal, $\hat{x}$ is the beamformer output signal, $||\cdot||_1$ is the L1-norm, and $L_{MR}$ is the multiresolution STFT loss. The multiresolution STFT loss uses a weight of 1 for the spectral convergence loss term, a weight of 1 for the log magnitude loss term, a weight of 4 for the linear magnitude loss term. It also uses Hanning windows with FFT sizes $[1024, 2048, 512]$, hop sizes $[120, 240, 50]$, and window lengths $[600,1200,240]$. The learning rate is initially 0.001 and we halve it if the loss function does not improve after 8 epochs.

\subsection{Beamformer real-world evaluation}\label{sec:beamformereval}
We evaluate the beamformer on real world recorded data from 6 unseen human participants in 3 unseen rooms. We group participants in pairs, and record data for every pair of participants in a different room. Each participant wears a microphone around each ear to record a binaural recording. The pair of participants take turns speaking for 8-10 minutes, with both participants recording audio the entire time. We process the recordings to slice out sections of self-speech recordings (same recorder and speaker) and interfering speech recordings (different recorder and speaker). Then, we create 100 5-second mixtures per speaker by combining a 5-second crop of self-speech and with a 5-second crop of interfering speech recorded by the same participant. We scale the power of each segment in a similar fashion as described in  \xref{sec:beamformerdatasets}, and then further scale the interfering speech so the scaled SNR is now uniformly sampled from $[-5, 5]$~dB. We report the results on this out-of-distribution beamformer dataset in Table~\ref{tab:bf_eval}, clearly showing significant noise reduction and self-speech extraction.

\begin{table}[t!]
\caption{Beamformer evaluation on unseen real-world mixtures. DNSMOS BAK is the estimate of the ITU P.835 background noise quality using a  neural net.}
\vskip -0.1in
\centering
{\footnotesize
\begin{tabular}{lccc}
\toprule
\textbf{Metrics} & SNR~(dB) & SI-SDR~(dB) & DNSMOS  \\
 &  & & BAK  \\
\midrule
    Mixture & -0.13 & -0.13 & 1.94 \\
    Beamformer & 8.36 & 7.78 & 3.96 \\
\bottomrule
\end{tabular}
}
\label{tab:bf_eval}
\vskip -0.15in
\end{table}

\section{Datasets}
We detail the dataset generation and spatialization process. With the exception of the Libri (leaving) dataset, the interference conversation in all datasets is always composed of exactly 2 speakers, with the target and interference conversations never sharing a common speaker.


\subsection{Dataset Generation}\label{sec:datasetgen}
\textbf{Libri}. This is a combination of 5 datasets -- Libri (2 spk), Libri (3 spk), {Libri (4 spk), Libri (5 spk), and} Libri (leaving) -- each of which consists of 60-second conversation mixtures between a target conversation and an interference conversation. These conversations are synthesized by populating speaker timestamps from one conversation dataset (RAMC~\cite{ramc}), with audio from another dataset LibriTTS~\cite{zen2019libritts}. Libri has 16,000 training samples, 2,600 validation samples, and {1000} test samples. Among the 1000 test samples, there are 200 samples for Libri (2 spk), 200 samples for Libri (3 spk), 200 samples for Libri (4 spk), 200 samples for Libri (5 spk), and 200 samples for Libri (leaving). The input SNR for the target conversation is sampled uniformly from –10 to 10 dB.

\textit{Libri (2 spk)}. The target conversations in this dataset have exactly 2 speakers. Since our model relies on self-speech to identify other speakers in the target conversation, we initially populate the timestamps with the self-speaker's audio for a minimum total duration of 5 seconds. Subsequently, for every remaining timestamp, we randomly populate it with speech from either the self speaker or the conversation partner. To prevent one speaker from dominating the conversation, we ensure the other target speaker meets a minimum utterance duration of 5 seconds. 

\vskip 0.02in\noindent\textit{Libri (3 spk)}. The target conversations in this dataset have exactly 3 speakers. Similar to the generation procedure for Libri (2 spk), we begin by ensuring the self-speaker speaks for the first 5 seconds. For all subsequent utterances, we randomly pick one speaker from the target conversation and insert their corresponding Libri audio into the utterance. Finally, we verify that each of the two conversation partners has at least one utterance exceeding 5 seconds in duration. The interference conversation in this mixture has 2 speakers.

\vskip 0.02in\noindent\textit{Libri (4 spk) and Libri (5 spk) (Evaluation only)}. {The target conversations in these two datasets contain exactly four and five speakers, respectively. The generation procedure is the same as in Libri (2 spk) and Libri (3 spk), except that for each utterance, we populate the audio with a randomly sampled speaker from a pool of 4 or 5 speakers. We ensure that the self-speaker speaks for the first 5 seconds, and that every other speaker has at least one utterance longer than 5 seconds to ensure their participation in the conversation.}

\vskip 0.02in\noindent\textit{Libri (leaving)}. Since human conversations are highly dynamic, our model must adapt to both preserve target speakers and suppress them when they leave the conversation and join the interference. To model this behavior, we generate a dataset that initially consists of a 3-speaker target conversation and a 2-speaker interference, which then transitions into a 2-speaker target conversation and a 3-speaker interference. We first ensure the self speaker speaks consecutively for at least 5 seconds at the start of the conversation. 

Then, one of the two conversation partners is chosen at random to leave the target conversation and join the interference conversation. Specifically, we randomly select a timestamp from the interference conversation that starts after that chosen timestamp for the conversation partner and before the 40 second mark to use for their first utterance in the interference conversation. We also require the leaving speaker to have a consecutive 5-second utterance in the interference conversation to confirm their presence. After this transition, the target conversation becomes a two-speaker conversation, and the interference becomes a three-speaker conversation.

\vskip 0.02in\noindent\textbf{Candor}. {This is a dataset consisting of 60-second conversation mixtures between a target and interference conversation from Candor. Since Candor dataset does not provide predefined splits, we created our own own by assigning 80\% of speakers to training, 10\% to validation, and 10\% to testing. Thus, we ensure there are no overlapping speakers across splits. When generating conversation mixture, we randomly select two recordings from the same split and ensure that they do not share any speakers. For the target conversation, we extract the 60-second segment where the self-speaker speaks continuously for at least 5 seconds at the beginning. In total, we generate 7000 training samples, 900 validation samples and 500 testing samples. Similar to the Libri datasets, the input SNR for the target conversation is sampled uniformly from –10 to 10~dB.}


\subsection{Dataset Spatialization}\label{sec:spatialize}

We generate synthetic egocentric audio using PyRoomAcoustics, an open-source room acoustics simulator widely used in audio research. The simulator produces left- and right-channel room impulse responses (RIRs) from each speaker, including the wearer, to microphones placed at the wearer’s ears.

Rooms have dimensions sampled uniformly: length and width from $[5, 10]$~m, and height from $[3, 4]$~m. The user is positioned at a distance uniformly sampled from $[0, 1]$m from the room center; other speakers are placed at distances from $[0.5, 1.5]$m. Person heights are sampled from $\mathcal{N}(175\text{cm}, 7\text{cm})$.

Microphones are placed near the user’s ears, offset laterally from the head center by half the head width, sampled from $\mathcal{N}(15~\text{cm}, 2~\text{cm})$. Audio sources are placed near each speaker’s mouth: vertically offset along the negative z-axis by $\mathcal{N}(18~\text{cm}, 2~\text{cm})$, and horizontally offset from their center by $\mathcal{N}(10.75~\text{cm}, 2~\text{cm})$.

Room reverberation time (RT60) is sampled uniformly from $[0.15, 1]$~s, capturing a range of acoustic environments.


\section{Training  details}\label{sec:trainingdetails}
In the first stage, we pretrain on 2K Libri (2spk) mixtures, 7K Libri (3spk) mixtures, 7K Libri (leaving) mixtures, and 7K Candor mixtures. The fast streaming model and the slow conversation embedding model are trained jointly without any pre-training. The two models were jointly trained for 120 epochs, with an initial learning rate of 0.002, AdamW optimizer with a weight decay of 0.01, and clip gradient norms to 1. We halve the learning rate if the loss does not decrease after 8 epochs. We use the negative SNR loss function and a batch size of 16 on 8 L40s.

{In the second stage, we jointly train the two models on the spatialized dataset (procedure outlined in Appendix~\ref{sec:spatialize}). The slow model is initialized with the pretrained weights from Stage 1, while the fast model is initialized from scratch. The models are jointly trained for 50 epochs with a with an initial learning rate of 0.002, AdamW optimizer with a weight decay of 0.01, and clip gradient norms to 1. We halve the learning rate if the loss does not decrease after 4 epochs. We use the negative SNR loss function and a batch size of 16 on 8 L40s.}

In the final stage, we augment our datasets by changing the duration of silence between every successive conversation partner utterance by a random amount sampled from $\mathcal{N}(0, 0.5~s)$. To preserve the order of utterances from the same speaker, we clip all silent durations to at least 1 sample. Finally, to retain the same overall duration of silence in the clip, we then normalize the length of each silent duration by sum of all silent duration in the audio clip. As a result, the speaker utterances occur at slightly different (and random) times, often overlapping with the user's speech. Here, we use the AdamW optimizer with a weight decay of 0.01, clip gradient norms to 1, and an initial learning rate of 0.0005. We halve the learning rate if the loss does not decrease after 8 epochs. This stage was trained for 42 epochs with a batch size of 4 on 2 A100 GPUs and the negative SNR loss function.

\begin{figure}[t!]
    \centering
    \includegraphics[width=\linewidth]{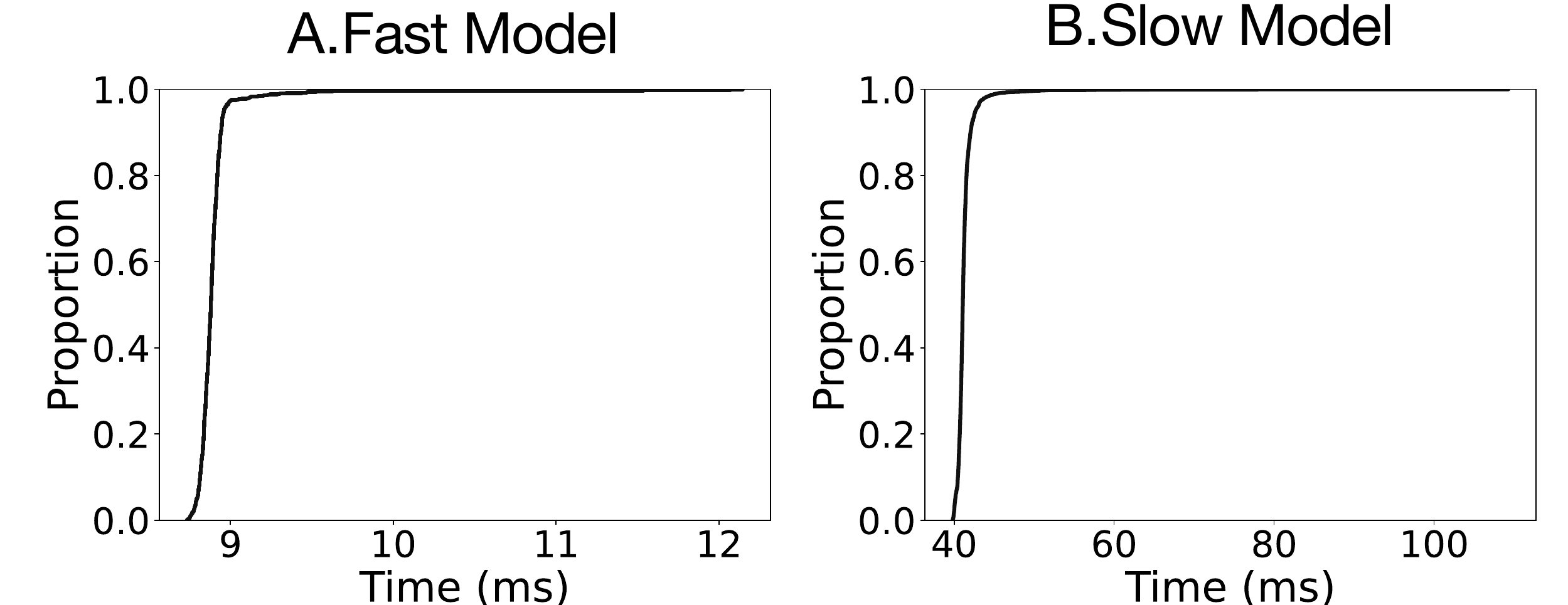}
    \vskip -0.05in
    \caption{Runtime of slow and fast model. A. The inference time of the fast model to process 12.5ms chunk on Orangpi 5B. B. The inference time of the slow model to process 1s chunk on Apple Silicon M2.}
    \vskip -0.15in
    \label{fig:runtime}
\end{figure}

\section{Runtime analysis}\label{sup:runtime}

Fig.~\ref{fig:runtime} reports CDF plots for the inference time of both the fast and slow models on the OrangePi 5B and Apple M2 silicon, respectively. 


\section{Conversation waveform examples}\label{sec:waveform}
{Fig. \ref{fig:waveform} shows an example from the spatialized Libri (2 spk) dataset. The mixture audio contains both the target and interference conversations. The beamformed self-speech is generated by applying our beamformer to this mixture. Due to spatialization, the self-speech is emitted closer to the wearer's ears, resulting in a higher amplitude compared to the other target speaker. However, our model was able to capture the low-amplitude speech of the other speaker. }

\section{Details of Ablation studies}\label{seC:ablationdetails}

\textit{Dual-model versus single model}. The fast only streaming model is trained on the same dataset as our dual model from stage 1, with a learning rate of 0.002, AdamW optimizer, a negative SNR loss function, and a batch size of 16 on 8 L40s. After training, we evaluate the model on the Libri 2-speaker testing dataset. 


\noindent\textit{Conversation embedding update rate}. The two dual-models are trained from scratch for 10 epochs using a learning rate of 0.001, the AdamW optimizer, a negative SNR loss, and a batch size of 8 on 8 RTX 6000 GPUs. After 10 epochs, the models show a consistent trend where the 1-second update rate outperforms the 4-second rate. Evaluation was conducted on 200 Libri 2-speaker test samples.

\begin{figure}[t]
    \centering
    \includegraphics[width=\linewidth]{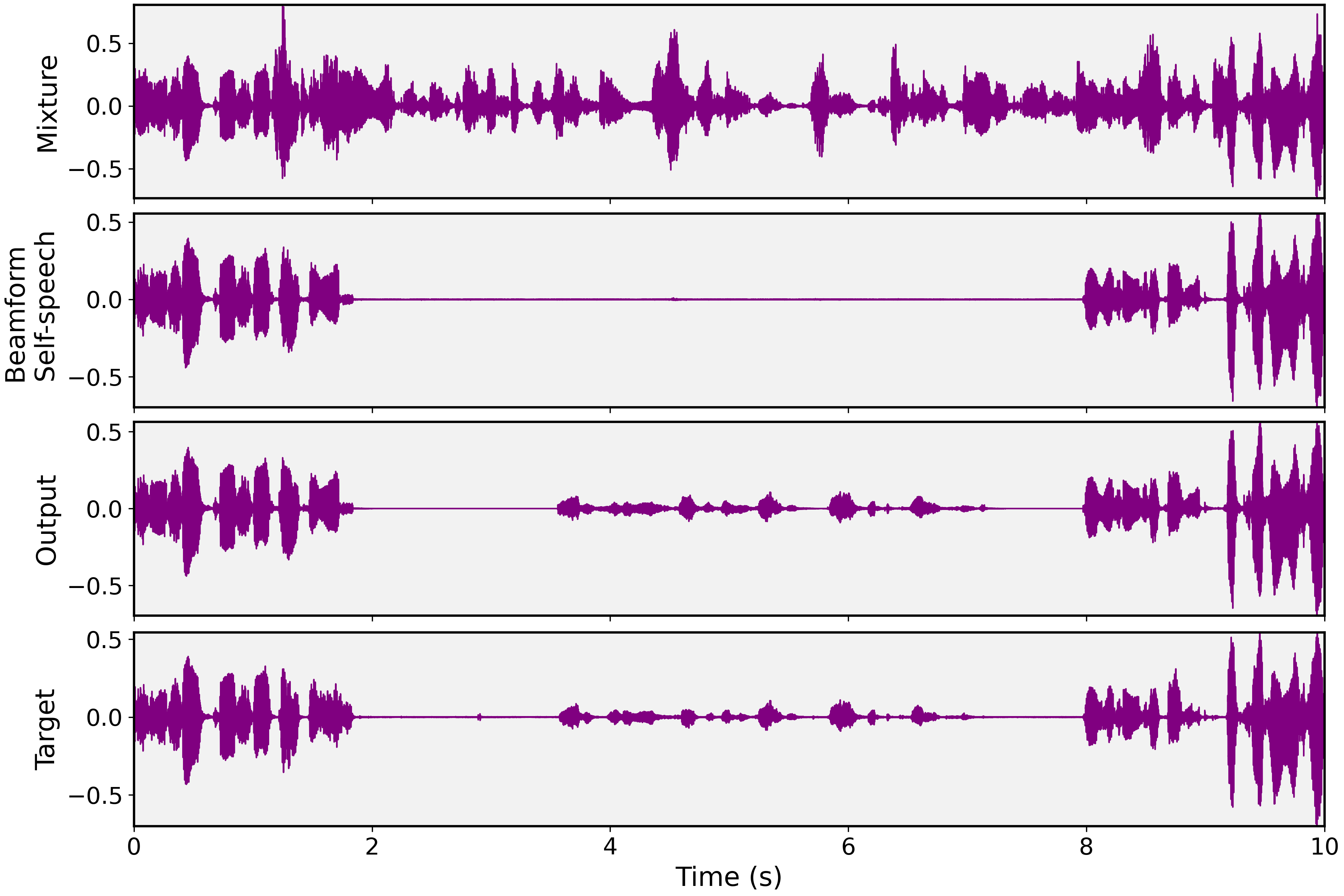}
    \vskip -0.1in
    \caption{Visualized waveforms for conversation mixture, the beamformed self-speech, output of the model and the groundtruth target egocentric conversation.}
    \label{fig:waveform}
\end{figure}

\noindent\textit{Speaker embedding versus self-speech}. We trained two slow models on the {Libri 2-speaker} training dataset. One model uses speaker embeddings and the other uses self-speech. Both models were trained for 100 epochs with a learning rate of 0.001, the AdamW optimizer, a negative SNR loss on 4 L40s. Evaluation was performed on 200 {Libri 2 speaker} test samples. 

\noindent\textit{Beamforming versus groundtruth self-speech}. For this ablation study, we used the model trained in stage 2  and evaluated it on 200 spatialized Libri 2-speaker test samples. During evaluation, the model was conditioned on either the original self-speech audio present in the mixture or the self-speech audio output by our beamformer.

\noindent\textit{Impact of turn-taking disruption} {We changed the duration of silence between two consecutive utterances from every speaker by sampling a duration shift from a normal distribution with mean 0 and a standard deviation parameter SD. This duration shift is added to the original silence duration and clipped  to preserve the order of the utterances. The lengths of the new silent sections are normalized so that the overall silence duration remains the same. With this perturbation technique, the parameter SD controls the extent to which the turn-taking dynamics are changed. Larger values of the standard deviation correspond to larger distortions of the original turn-taking structure. This process maintains the overall speech content but disrupts the natural temporal flow of the conversation. We then evaluate the performance of our model trained in stage 1 on our Candor test set with different  perturbations.
}

\noindent\textit{Context length.}  {We trained 4 dual models with different context length configurations (1s, 5s, 10s, full context) for 20 epochs with a learning rate of 0.001, AdamW optimizer and negative SNR loss. We train the models on Libri 2-speaker training set and evaluate them on 200 Libri 2 speaker test samples.}

\begin{table}[t!]
\caption{Statistics for our real egocentric conversation recordings. }
\vskip -0.1in
\centering
{\footnotesize
\begin{tabular}{lr}
\toprule
 Statistic &  Mean (STD)\\
\midrule
Turn-change Frequency ($min^{-1}$) & 6.2 (4.6)  \\
Turn Duration &  8.2s (8.8s)    \\
Overlap Ratio &  1.3\% (2.5\%)    \\
IPU Duration ($min^{-1}$) & 52.0s (3.54s) \\
FTO & 0.18s (1.38s)  \\
\bottomrule
\end{tabular}
}
\label{tab:realworldstats}
\vskip -0.15in
\end{table}

\section{{User Study Design}}\label{sec:user_study}
{We ask each participant the following four questions after they listen to both the original audio mixture and our model's output. Each question used a 5-point scale.}

\begin{enumerate}[label=(\arabic*), leftmargin=*, itemsep=0pt, topsep=0pt]
  \item \textbf{Noise Suppression}: How INTRUSIVE/NOTICEABLE were the INTERFERING SPEAKERS? 5 - Not noticeable; 4 - Slightly noticeable; 3 - Noticeable, but not intrusive; 2 - Somewhat intrusive; 1- Very intrusive

  \item \textbf{Conversation Comprehension}: How EASY was it to understand the target conversation in this audio sample? 5 - Very easy; 4 - Easy; 3 - Neutral or Neither easy nor hard; 2 - Hard; 1 - Very hard

  \item \textbf{Effort}: How much EFFORT did it take to focus on the target conversation in this audio sample? 5 - Very little effort; 4 - Little effort; 3 - Moderate effort; 2 - High effort; 1 - Very high effort

  \item \textbf{Overall MOS}: If the goal is to focus on this target conversation, how was your OVERALL experience? 5 - Excellent; 4 - Good; 3 - Fair; 2 - Poor; 1 - Bad
\end{enumerate}

\section{Egocentric evaluation participants}
The study was approved by our institution's IRB. All participants provided informed consent and were recruited from our institution and nearby areas. They were offered a \$15 compensation. 

\section{Real-egocentric conversation analysis}

We compute several conversational statistics from our collected real-world egocentric recordings:
\squishlist
\item {\it Turn-Change Frequency.} The number of speaker turn changes per minute.

 \item {\it Turn Duration.} The length of each individual speaking turn.

 \item  {\it Overlap Ratio.} The proportion of time during which multiple speakers talk simultaneously.

 \item {\it  Interpausal Unit (IPU) Duration.} A continuous stretch of speech from a single speaker, bounded by silences longer than 200 ms on both sides, as detected by a voice activity detector.

 \item  {\it  Floor-Transfer Offset (FTO).} The time gap between the end of one speaker’s turn and the start of the next, which is a combination of overlaps and gaps. Negative values indicate overlapping speech, while positive values indicate gaps between turns.

\squishend


\end{document}